# A HYBRID MODEL FOR BANKRUPTCY PREDICTION USING GENETIC ALGORITHM, FUZZY C-MEANS AND MARS

[1]A.Martin [2]V.Gayathri [3]G.Saranya [4]P.Gayathri [5]Dr.Prasanna Venkatesan

**[1]**Research scholar, Dept. of Banking Technology, Pondicherry University, Pondicherry, India.

jayamartin@yahoo.com

**[2,3,4]**Student of IT, Sri Manakula Vinayagar Engineering College, Pondicherry, India.
[2]gayuvel@gmail.com [3]saranyas228@gmail.com [4]gayatripanneer.it@gmail.com

**[5]**Associate Professor, Dept. of Banking Technology, Pondicherry University, Pondicherry, India.
prasanna_v@yahoo.com

## ABSTRACT

*Bankruptcy prediction is very important for all the organization since it affects the economy and rise many social problems with high costs. There are large number of techniques have been developed to predict the bankruptcy, which helps the decision makers such as investors and financial analysts. One of the bankruptcy prediction models is the hybrid model using Fuzzy C-means clustering and MARS, which uses static ratios taken from the bank financial statements for prediction, which has its own theoretical advantages. The performance of existing bankruptcy model can be improved by selecting the best features dynamically depend on the nature of the firm. This dynamic selection can be accomplished by Genetic Algorithm and it improves the performance of prediction model. .*

## 1. INTRODUCTION

Bankruptcy prediction remains a concern for the various stakeholders in a firm, owners, managers, investors, creditors and business partners, as well as government institutions responsible for maintaining the stability of financial markets and general economic prosperity. Analysis of the financial position of the firm is very useful for the banking and financial industry, such a task for credit risk assessment. The insight of both practitioners and researchers, originating with the papers of Beaver (1967) [12] and Altman (1968) [13], was that firms with certain financial structures have a greater probability of default, and eventual bankruptcy, than other types of firms.

 

Many models have been proposed then; particularly key cash flow and debt ratios appear to be important predictors of forth coming bankruptcy. Some of these proposals have been applied by lending institutions for credit scoring purposes. These models have quite heterogeneous characteristics but most of them share the common feature of relying on multivariate techniques whose input variables are mainly financial descriptors of the credit applicant. Alternative approaches have developed scoring models based on market information and group decision analysis. However, the alternative approaches do not significantly outperform models based on multivariate techniques. During recent years researchers have paid special attention to non-parametrical and hybrid approaches. Non-parametrical techniques are suitable for bankruptcy prediction tasks due to the specific features of financial information (i.e. non-normality and heteroskedasticity). Hybrid approaches combine several classification methods and achieve greater accuracy than individual models. In the present research we propose a strategy for the construction of a hybrid model which is based on the sequential application of genetic algorithm, fuzzy c-means clustering and Multivariate Adaptive Regression Splines (MARS) [6]. The classification nature of these techniques is very much suitable for the field of finance. Most of the approaches that exist use a particular financial ratio model commonly like Altman, Ohlson, Zmijewski, etc. Using Genetic algorithm we can choose the financial ratio set dynamically by considering features such as accuracy in prediction, suitability of bankruptcy model for an industry, Number of ratios used for prediction, suitability of classification model and  that can increase the accuracy of the prediction. An advantage of Genetic Algorithm is that it is capable of extracting optimal rules that can be integrated to any system [3]. There are many features for a firm to undergo bankruptcy. Type I and Type II errors are the errors of misclassifying a bankrupt firm as healthy, and misclassifying a healthy firm as bankrupt, respectively [7].

## 2. PRIOR RESEARCH

There are many Classification techniques are used to predict bankruptcy. These classification techniques are broadly classified into single classification and hybrid model.

### 2.1 Single Classification Model

#### 2.1.1 Fuzzy Rule Based Model

The earliest applications of fuzzy set theory are knowledge based systems whose core is a set of fuzzy `if-then' rules [9]. Most of these systems derive these fuzzy `if-then' rules from human experts. The generation of fuzzy if-then rules from numerical data involves (i) the fuzzy partition of a pattern space into fuzzy subspaces and (ii) the determination of fuzzy if-then rules for each fuzzy partition [9, 17]. The performance of such a classification system depends on the choice of a fuzzy partition. The fuzzy classifier produced average classification rate of 88.3%, 33.3%, 79.2% and 89.2% respectively for two, three, four and five partitions of the input space. Even though, the case of five partitions yielded highest average classification rate, the average Type-I error is higher compared to the case of two partitions. The average Type I error for two partitions is zero, which is a remarkable result. This means that all the bankrupt banks are classified correctly [8].

### 2.1.2 Neural Network

Back-Propagation Neural Network model is the most popular methodology in the neural networking. This model considers corporate factors and different sample matches but also uses financial ratios within different periods to construct Back-Propagation Neural Network models.

There are three layers in neural network, the first layer—input layer, the second—hidden layer and the third layer—output layer. The hidden layer can be one or more layers, but we adopted the typical three-layer architecture in this study. The transfer function was sigmoid function [5] as

$$f(t)=1/(1+e-t).$$

The neurons of input layer were predictors. The outputs of neurons in output layer were the output of BPNN. There are links to connect the neurons of each layer. Every link has a relative weight to represent the importance of input information. The algorithm of BPNN is to input data from input layer into hidden layer where data were transferred by the log sigmoid function, and then output into output layer. BPNN uses sample data for its prediction taken from open market of listed companies in Taiwan electronic industries for the effect of prediction correctness for one-year prior of crisis [5].

Training data selection and Building BPNN model— design an experiment with three different periods, one-year, two-year and three-year prior crisis, to test which period has the lowest type I error and type II error at the same time. The experimental results presented to apply health prediction pattern and crisis prediction pattern on a company which may not know its financial tendency (health or crisis) can get a 73.67% correctness future prediction if it had predicted outcome [5].

## 2.2 Hybrid Model

### 2.2.1 Neuro-Fuzzy Model

The process of making credit-risk evaluation decision is complex and unstructured. Neural networks are known to perform reasonably well compared to alternate methods for this problem. However, a drawback of using neural networks for credit-risk evaluation decision is that once a decision is made, it is extremely difficult to explain the rationale behind that decision. Researchers have developed methods using neural network to extract rules, which are then used to explain the reasoning behind a given neural network output. These rules do not capture the learned knowledge well enough. Neuro-fuzzy systems have been recently developed utilizing the desirable properties of both fuzzy systems as well as neural networks. These Neuro-fuzzy systems can be used to develop fuzzy rules naturally. In this model, we analyze the beneficial aspects of using both Neuro-fuzzy systems as well as neural networks for credit-risk evaluation decisions [10].

As a result, Neural networks performed somewhat better than Neuro-fuzzy systems in terms of classification accuracy, on both training as well as testing of data. This result is not surprising given the various approximations that are made while dealing with fuzzification / defuzzification and also the approximations that are made in fuzzy arithmetic, while learning the rules in the Neuro-fuzzy system [10].

### 2.2.2 Genetic, Fuzzy and Neural networks Model

Neural network is good for non-linear data with learning capacity. Its disadvantage is black box approach (after training difficult to trace how they work).Fuzzy Neural Network adds rules to Neural Network that overcome black box but learning capacity is reduced. Genetic Fuzzy Neural Network is proposed to avoid black box approach and to improve learning efficiency [4].

The total misclassification rate of NNs model is 4.17%, while the total misclassification rate of GFNN model is 1.67%. Specially, the type 1 error rate of the NNs model is 10.00%, while the type 1 error rate of GFNN is only 3.33%. It is clearly shown that the accuracy of bankruptcy prediction of GFNN model is much better than the NNs model [4].

## 3. OUR APPROACH

A strategy has been proposed for the construction of a hybrid model which is based on the sequential application of fuzzy clustering and Multivariate Adaptive Regression Spines (MARS). The reason for the choice of these techniques is that they are suited to the field of finance. Firstly, the use of clustering techniques is motivated by the existence of different failing processes. As failing firms may have dissimilar financial features. Secondly, MARS is a flexible procedure, which models relationships that are nearly additive or involve interactions with fewer variables. As a small number of latent factors are capable of representing a great percentage of the information contained in the annual accounts of firms, MARS is also suitable for the bankruptcy prediction problem [6].

Most of the bankruptcy prediction techniques use static financial ratios models for its prediction which will not use for all type of firms since most of the firms have different attributes. So these techniques will not be suitable for all types of firms. In our proposed work, we use Genetic algorithm to select an optimal model dynamically for the prediction which will improve the performance. The selected model is used to cluster the data using Fuzzy c-means Clustering and regression technique MARS is used to predict whether bankrupt or not.

### 3.1 Genetic Algorithm

Genetic Algorithms are general-purpose search and optimization procedures. They are inspired by the biological evolution principle of survival of the fittest. The genetic algorithm (GA) is a search heuristic that mimics the process of natural evolution. This heuristic is routinely used to generate useful solutions to optimization and search problems. Genetic algorithms belong to the larger class of evolutionary algorithms (EA), which generate solutions to optimization problems using techniques inspired by natural evolution, such as inheritance, mutation, selection, and crossover in Figure 4.

Figure 4. Steps in Genetic Algorithm to find best optimal solution [15]

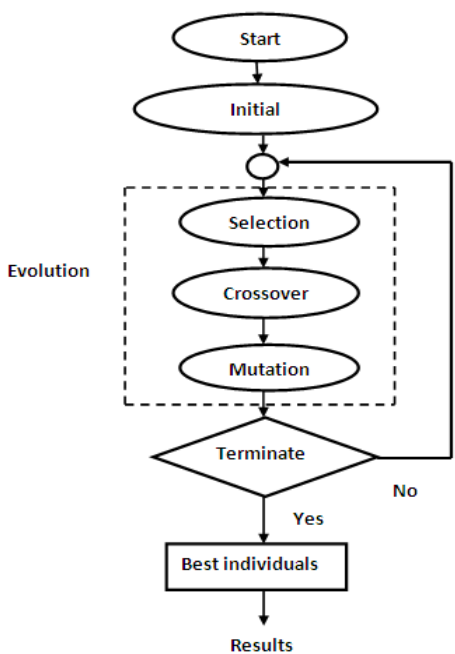

GAs does not deal directly with the parameters of the problem to be solved. They work with codes which represent the problem and produce codes which represent the solution. A typical genetic algorithm requires:

1. A genetic representation of the solution domain,

2. A fitness function to evaluate the solution domain.

3. GAs were reported to have achieved great success because of their ability to exploit accumulated information about an initially unknown search space leading to moving subsequent searches into more useful subspaces. GAs is also flexible tools and can be used in combination with other techniques including fuzzy logic and neural networks [15].

### 3.2 Fuzzy c-means Clustering

The Fuzzy c-means algorithm is a method of clustering which allows one piece of data to belong to two or more clusters. This method was developed by Dunn (1973) [2] and improved by Bezdek (1981) [18]. It is frequently used in pattern recognition. Under the Fuzzy c-means approach, each given datum does not belong exclusively to a well defined cluster, but it can be placed in a middle way. This belonging to more than one cluster is represented by probability coefficients. This method is based on minimization of the following objective function

$$J = \sum_{i=1}^{N} \sum_{j=1}^{C} u_{ij}.\left\| x_i - c_j \right\|^2$$

Where $u_{ij}$ is the probability of membership of the company $x_i$ to the jth cluster; $x_i$, the ith company; $c_j$, the coordinates of the jth center of the cluster; $\|\ \|$ represents the Euclid norm; N, the number of companies in the database; C is the total number of clusters considered. Fuzzy partitioning is carried out by an iterative optimization of the objective function with the update of membership $u_{ij}$ and the cluster centers $c_j$, by:

$$u_{ij} = \frac{1}{\sum_{k-1}^{C} \left( \frac{\left\| x_i - c_j \right\|}{\left\| x_i - c_k \right\|} \right)^2}$$

$$c_j = \frac{\sum_{i=1}^{N} u_{ij} \cdot x_i}{\sum_{i=1}^{N} u_{ij}}$$

As already mentioned, in the Fuzzy c-means algorithm, each given datum does not belong to a well defined cluster, but it can be placed in a middle way. This belonging to more than one cluster is represented by the probability coefficients. So, the company $x_i$ has a probability $u_{ij}$ of belonging to cluster j, $\sum_{j=1}^{C} u_{ij} = 1$

### 3.3 Multivariate Adaptive Regression Splines (MARS) model

As stated earlier Multivariate Adaptive Regression Splines (MARS) is a multivariate nonparametric regression technique developed by Friedman (1991) [17]. Its main purpose is to predict the values of a continuous dependent variable, $\vec{y} = (n \times 1)$ from a set of independent explanatory variables, $\vec{X}(n \times p)$. The MARS model can be represented as:

$$\vec{y} = f\left( \vec{X} \right) + \vec{e}$$

where $\vec{e}$ is an error vector of dimension $(n \times 1)$.

MARS can be considered as a generalisation of classification and regression trees (CART) (Hastie, Tibshirani, & Friedman, 2003) [19], and is able to overcome some limitations of CART. MARS does not require any priori assumptions about the underlying functional relationship between dependent and independent variables. Instead, this relation is covered from a set of coefficients and piecewise polynomials of degree q (basis functions) that are entirely driven from the regression data $(\vec{X}.\vec{y})$.

17

# 4. DYNAMIC FEATURE SELECTION USING GENETIC ALGORITHM

## 4.1 Various Feature models

For the bankruptcy prediction, there are 5 models used popularly for all the techniques. In this model, we are going to use all the 5 models as different features and going to select the best prediction model by using Genetic Algorithm [1].

### 4.1.1 Altman Model

A predictive model created by Edward Altman in the 1960s. This model combines five different financial ratios to determine the likelihood of bankruptcy amongst companies. The Altman [1968] [12] study is used as the standard for comparison for subsequent bankruptcy classification studies using discriminant analysis. The ratios are:

$$Z = 0.012X_1 + 0.014X_2 + 0.033X_3 + 0.006X_4 + 0.999X_5.$$

Where
*X1 = Net working capital/total assets*
*X2 = Retained earnings/total assets*
*X3 = Earnings before interest and taxes/total assets.*
*X4 = Market value of equity/book value of total liabilities.*
*X5 = Sales/total assets.*

Figure 5. Set A Altman Model

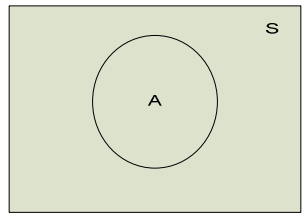

where Set A represents the Altman Model

### 4.1.2 Ohlson Model

Ohlson employed logistic regression to predict company failure [10], to overcome the limitations of Altman. This model consists of the following ratios:

$Z = 1/ [1 exp - (a [b.sub.1] TLTA [b.sub.2] WCTA [b.sub.3] CLCA [b.sub.4] OENEG [b.sub.5] NITA [b.sub.6] FUTL [b.sub.7] INTWO [b.sub.8] CHIN)]$

Where,

*Z = the probability of bankruptcy for a firm*
*TLTA = Total liabilities divided by total assets.*
*WCTA = Working capital divided by total assets.*
*CLCA = Current liabilities divided by current assets.*
*OENEG = 1 If total liabilities exceed total assets, 0 otherwise.*
*NITA = Net income divided by total assets.*
*FUTL = Funds provided by operations (income from operation after depreciation) divided by total liabilities.*
*INTWO = 1 If net income was negative for the last 2 years, 0 otherwise.*
*CHIN = (NIt _ NIt_1)/ (|NIt| + |NIt_1|), where NIt is net income for the most recent period. The denominator acts as a level indicator. The variable is thus intended to measure the relative change in net income.*

Figure 6. Set B Ohlson Model

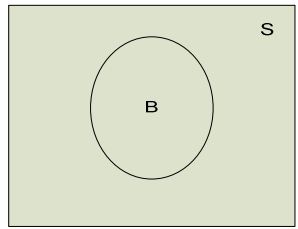

where Set B represents the Ohlson Model

### 4.1.3 Zmijeswski Model

This model is used to predict bankruptcy using probit model [15]. It consists of 3 ratios as:

*NITL = Net income divided / total liabilities.*
*TLTA = Total liabilities divided / total assets.*
*CACL = Current assets divided / current liabilities.*

Figure 7. Set C Zmijeswski Model

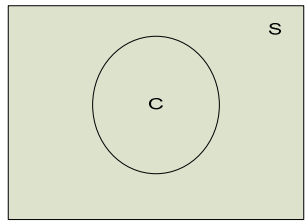

where Set C represents the Zmijeswki Model

### 4.1.4 Shumway Model

Shumway improved on the basic bankruptcy models by combining market ratios along with the traditional financial ratios [13]. This model is defined as follows:

$$Z = 1/ [1 \ exp - (a \ [b.sub.1] \ NITA \ [b.sub.2] \ TLTA \ [b.sub.3] \ ERR \ [b.sub.4] \ SDR)]$$

*Where*
*Z = the probability of bankruptcy for a firm*
*NITA = Net income/total assets*
*TLTA = Total liabilities/total assets*
*ERR = Excess rate of return (i.e., a firm's rate of return minus the market's rate of return)*
*SDR = Standard deviation of residual returns (Residual return = a firm's realized rate of return - its expected rate of return)*

Figure 8.  Set D Shumway Model

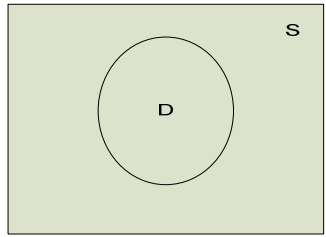

where Set D represents the Shumway Model

Table 1 Summary of empirical models and variables employed

| Model | Variable | Ratios |
|---|---|---|
| Multiple – Discriminant analysis<br><br>Altman (1968) | X1 | Net working capital / total assets |
|  | X2 | Retained earnings/total assets |
|  | X3 | Earnings before interest and taxes/total assets |
|  | X4 | Market value of equity/book value of total liabilities. |
|  | X5 | Sales/total assets |
| Logit model<br><br>Ohlson (1980) | TLTA | Total liabilities divided by total assets |
|  | WCTA | Working capital divided by total assets |
|  | CLCA | Current liabilities divided by current assets |
|  | NITA | Net income divided by total assets |
|  | FUTL | Funds provided by operations (income from operation after depreciation) divided by total liabilities |
| Probit model | NITL | Net income divided by total liabilities. |

| Zmijewski (1984) | TLTA | Total liabilities divided by total assets. |
|---|---|---|
| | CACL | Current assets divided by current liabilities. |
| Hazard model Shumway (2001) | NITL | Net income divided by total liabilities. |
| | TLTA | Total liabilities divided by total assets. |

## 4.3 Results

### 4.3.1 Database

 The data sets are taken with the help of Centre for Monitoring Indian Economy Pvt. Ltd. The Centre for Monitoring Indian Economy is an independent economic think-tank headquartered in Mumbai, India. It provides information solutions in the form of databases and research reports. CMIE has built the largest database on the Indian economy and companies. Prowess is a database of large and medium Indian firms. It contains detailed information on over 25,346 firms. From that software we were taken three types of banks as successful banks, recovered banks and failed banks as a training data.

The data in Microsoft Excel Sheet is stored according to the models mentioned above, by referring their respective ratios. The various feature models are analyzed by comparing their ratios and by using Genetic Algorithm, the models which have same ratio that of others will form another model. From the Fig.9 we have created a new set E which is the union of set B, C and D.

Figure 9. Set E Forming a new model

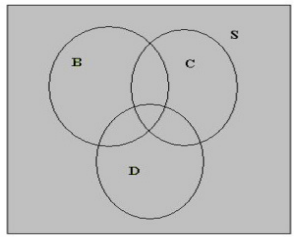

where Set E represents the new model which is the optimal model for predicting bankruptcy formed using Genetic algorithm.

The intersection of set E consists of a ratio which is common to the other sets as B, C & D. Using this new set, Genetic Algorithm predicts the bankruptcy with higher performance than the other classification models.

By using genetic algorithm the best model can be chosen and that result is used to create cluster by using Fuzzy c-means algorithm using MATLAB7.0. The result will be as like as in the Figure 10 with three clusters as Bankrupt, non bankrupt and may/may not bankrupt.

Figure 10. Data Clustering for Bankruptcy Prediction models

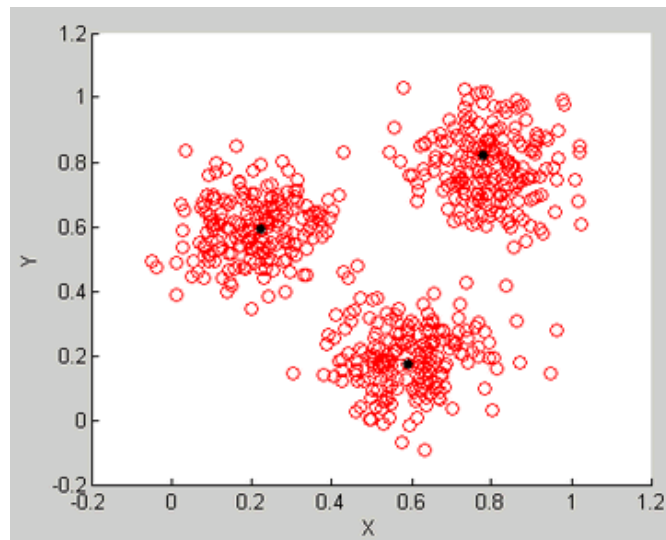

From these clusters, finally MARS model is used to classify the organization as bankrupt or not bankrupt from the test data.

# 5. CONCLUSION

This paper conceptualizes the hybrid model of genetic algorithm, Fuzzy c-means algorithm, MARS for prediction of bankruptcy. It also addresses the issues in the fuzzy cluster creation for bankruptcy prediction models and also provides the solution using genetic algorithm to select the bankruptcy model dynamically. In this model, Genetic algorithm selects the best ratio set by considering the various factors of the given firm, so the resultant cluster formed is good. Since MARS can be considered as a Classification techniques, it does not require functional relationship between dependent and independent variable. The results of fuzzy clustering can be integrated with MARS. This hybrid model is performs well compared to static Bankruptcy Models.

**Authors**

Mr.A.Martin Assistant Professor in the Department of Information Technology in Sri Manakula Vinayagar Engineering College. Pondicherry University, Pudhucherry, India. He holds a M.E and pursuing his Ph.D in Banking Technology from Pondicherry University, India. He can be reached via jayamartin@yahoo.com

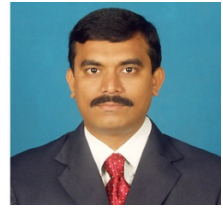

Ms.V.Gayathri, Student of Information Technology, Sri Manakula Vinayagar Engineering College. Pondicherry University, Pudhucherry. She can be reached via gayuvel@gmail.com.

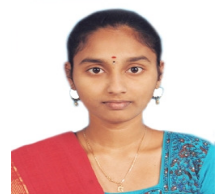

Ms.G.Saranya, Student of Information Technology, Sri Manakula Vinayagar Engineering College. Pondicherry University, Pudhucherry. She can be reached via saranyas228@gmail.com.

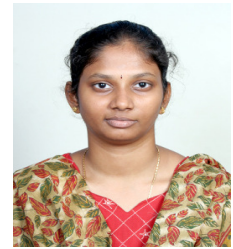

Ms.P.Gayathri, Student of Information Technology, Sri Manakula Vinayagar Engineering College. Pondicherry University,Pudhucherry. She can be reached via gayatripanneer.it@gmail.com.

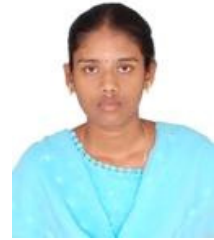

Dr.Prasanna Venkatesan, Associate Professor, Dept. of Banking Technology, Pondicherry University, Pondicherry. He has more than 20 years Teaching and Research experience in the field of Computer Science and Engineering; He has developed a compiler for multilingual languages, he can be reached via prasanna_v@yahoo.com.

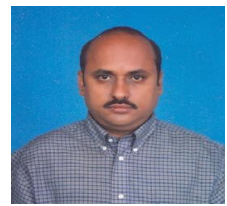